\documentclass{svmult}
\usepackage{mathptmx}
\usepackage{helvet} 
\usepackage{courier}
\usepackage{type1cm}
\usepackage{bigstrut}
\usepackage{makeidx}
\usepackage{graphicx} 
\usepackage{multicol} 
\usepackage[bottom]{footmisc}
\usepackage{multirow}
\usepackage{amsmath,amssymb,amsfonts}

\newcommand{\makefigure}[2]{
	\begin{figure}[h]
		\begin{center}
			\includegraphics[width=120mm]{#1}
		\end{center}
		\caption{#2}
		\label{fig:#1}
	\end{figure}
}

\DeclareMathOperator*{\logadd}{logadd}

\makeindex   

\begin{document}

\title*{Sentiment Analysis, Basic Tasks of}
\author{Iti Chaturvedi, Soujanya Poria, Erik Cambria}
\institute{Iti Chaturvedi, Soujanya Poria, Erik Cambria \at School of Computer Science and Engineering,\\Nanyang Technological University, Singapore\\\email{\{iti, sporia, cambria\}@ntu.edu.sg}}
%\institute{Soujanya Poria \at Temasek Laboratory, \\Nanyang Technological University, Singapore \\\email{sporia@ntu.edu.sg}}

\maketitle

\section{Affiliation}

School of Computer Science and Engineering, Nanyang Technological University, Singapore

\subsection{Synonyms}

Sentiment Analysis, Subjectivity Detection, Deep Learning Aspect Extraction, Polarity Distribution, Convolutional Neural Network.

\subsection{Glossary}

Aspect : Feature related to an opinion target\\
Convolution : features made of consecutive words\\
BOW : Bag of Words\\
NLP : Natural Language Processing\\
CNN : Convolutional Neural Network\\
LDA : Latent Dirichlet Allocation\\

\subsection{Definition}

Subjectivity detection is the task of identifying objective and subjective sentences. Objective sentences are those which do not exhibit any sentiment. So, it is desired for a sentiment analysis engine to find and separate the objective sentences for further analysis e.g., polarity detection. In subjective sentences, opinions can often be expressed on one or multiple topics. Aspect extraction is a subtask of sentiment analysis that consists in identifying opinion targets in opinionated text, i.e., in detecting the specific aspects of a product or service the opinion holder is either praising or complaining about.

\subsection{Key Points}

\begin{itemize}
\item
We consider deep convolutional neural networks where each layer is learned independent of the others resulting in low 
complexity.

\item
We model temporal dynamics in product reviews by pre-training the deep CNN using dynamic Gaussian Bayesian networks.

\item 
We combine linguistic aspect mining with CNN features for effective sentiment detection.

\end{itemize}

\subsection{Historical Background}

Traditional methods prior to 2001 used hand-crafted templates to identify subjectivity and did not generalize well for resource-deficient languages such as Spanish. Later works published between 2002 and 2009 proposed the use of deep neural networks to automatically learn a dictionary of features (in the form of convolution kernels) that is portable to new languages. Recently, recurrent deep neural networks are being used to model alternating subjective and objective sentences within a single review. Such networks are difficult to train for a large vocabulary of words due to the problem of vanishing gradients. Hence, in this chapter we consider use of heuristics to learn dynamic Gaussian networks to select significant word dependencies between sentences in a single review. 

Further, in order to relation between opinion targets and the corresponding polarity in a review, aspect based opinion mining is used. Explicit aspects were models by several authors using statistical observations such mutual information between noun phrase and the product class. However this method was unable to detect implicit aspects due to high level of noise in the data. Hence, topic modeling was widely used to extract and group aspects, where the latent variable 'topic' is introduced between the observed variables 'document' and 'word'. In this chapter, we demonstrate the use of 'common sense reasoning' when computing word distributions that enable shifting from a syntactic word model to a semantic concept model. 

\section{Introduction}

While sentiment analysis research has become very popular in the past ten years, most companies and researchers still approach it simply as a polarity detection problem. In reality, sentiment analysis is a `suitcase problem' that requires tackling many natural language processing (NLP) subtasks, including microtext analysis, sarcasm detection, anaphora resolution, subjectivity detection and aspect extraction. In this chapter, we focus on the last two subtasks as they are key for ensuring a minimum level of accuracy in the detection of polarity from social media.

The two basic issues associated with sentiment analysis on the Web, in fact, are that (1) a lot of factual or non-opinionated information needs to be filtered out and (2) opinions are most times about different aspects of the same product or service rather than on the whole item and reviewers tend to praise some and criticize others. Subjectivity detection, hence, ensures that factual information is filtered out and only opinionated information is passed on to the polarity classifier and aspect extraction enables the correct distribution of polarity among the different features of the opinion target (in stead of having one unique, averaged polarity assigned to it). In this chapter, we offer some insights about each task and apply an ensemble of deep learning and linguistics to tackle both.

The opportunity to capture the opinion of the general public about social events, political movements, company strategies, marketing campaigns, and product preferences has raised increasing interest of both the scientific community (because of the exciting open challenges) and the business world (because of the remarkable benefits for marketing and financial market prediction). 
Today, sentiment analysis research has its applications in several different scenarios. 
There are a good number of companies, both large- and small-scale, that focus on the analysis of opinions and sentiments as part of their mission \cite{camacsa}.
Opinion mining techniques can be used for the creation and automated upkeep of review and opinion aggregation websites, in which opinions are continuously gathered from the Web and not restricted to just product reviews, but also to broader topics such as political issues and brand perception. 
Sentiment analysis also has a great potential as a sub-component technology for other systems. It can enhance the capabilities of customer relationship management and recommendation systems; for example, allowing users to find out which features customers are particularly interested in or to exclude items that have received overtly negative feedback from recommendation lists.
Similarly, it can be used in social communication for troll filtering and to enhance anti-spam systems. 
Business intelligence is also one of the main factors behind corporate interest in the
field of sentiment analysis \cite{cambig}.

Sentiment analysis is a `suitcase' research problem that requires tackling many NLP sub-tasks, including semantic parsing \cite{dhegra}, named entity recognition \cite{maalab}, sarcasm detection \cite{porloo}, subjectivity detection and aspect extraction.
In opinion mining, different levels of analysis granularity have been proposed, each one having its own
advantages and drawbacks \cite{camint,cambio}. Aspect-based opinion mining~\cite{Hu2004,XiaowenDing2008}
focuses on the relations between aspects and document polarity.
An aspect, also known as an opinion target, is a concept in which the opinion is expressed in the given document.
For example, in the sentence, ``The screen of my phone is really nice and its resolution is superb'' 
for a phone review contains positive polarity, i.e., the author likes the phone. However, 
more specifically, the positive opinion is about its \emph{screen} and \emph{resolution}; 
these concepts are thus called opinion targets, or aspects, of this opinion. 
The task of identifying the aspects in a given opinionated text is called aspect extraction.
There are two types of aspects defined in aspect-based opinion mining:
explicit aspects and implicit aspects. Explicit aspects are words
in the opinionated document that explicitly denote the opinion target.
For instance, in the above example, the opinion targets \emph{screen} and \emph{resolution} are explicitly mentioned in the text.
In contrast, an implicit aspect is a concept that represents the opinion target of an opinionated
document but which is not specified explicitly in the text. One can infer
that the sentence, ``This camera is sleek and very affordable'' implicitly contains
a positive opinion of the aspects \emph{appearance} and \emph{price} of the entity \emph{camera}.
These same aspects would be explicit in an equivalent sentence: 
``The appearance of this camera is sleek and its price is very affordable.''

Most of the previous works in aspect term extraction have either used conditional random fields (CRFs) \cite{jakob2010extracting,zhiqiang2014dlirec} or linguistic patterns \cite{Hu2004,porsent}. Both of these approaches have their own limitations: CRF is a linear model, so it needs a large number of features to work well; linguistic patterns need to be crafted by hand, and they crucially depend on the grammatical accuracy of the sentences.
In this chapter, we apply an ensemble of deep learning and linguistics to tackle both the problem of aspect extraction and subjectivity detection. 

The remainder of this chapter is organized as follows: Section~\ref{ssoa} and \ref{asoa} propose some introductory explanation and some literature for the tasks of subjectivity detection and aspect extraction, respectively; Section~\ref{prep} illustrates the basic concepts of deep learning adopted in this work; Section~\ref{algo} describes in detail the proposed algorithm; Section~\ref{eval} shows evaluation results; finally, Section~\ref{end} concludes the chapter.

\section{Subjectivity detection}
\label{ssoa}
Subjectivity detection is an important subtask of sentiment analysis that can prevent a sentiment classifier from considering irrelevant or potentially misleading text in online social platforms such as Twitter and Facebook. Subjective extraction can reduce the amount of review data to only 60$\%$ and still produce the same polarity results as full text classification \cite{Bonzanini:2012}. This allows analysts in government, commercial and political domains who need to determine the response of people to different crisis events \cite{Bonzanini:2012,NLE:8280994,RSM:Duy2014}. 
Similarly, online reviews need to be summarized in a manner that allows comparison of opinions, so that a user can clearly see the advantages and weaknesses of each product merely with a single glance, both in unimodal \cite{RSM:Hui2009} and multimodal \cite{porcon,camble} contexts. Further, we can do in-depth opinion assessment, such as finding reasons or aspects \cite{porasp} in opinion-bearing texts. For example, $\textrm{`Poor acting'}$, which makes the film $\textrm{`awful'}$. Several works have explored sentiment composition through careful engineering of features or polarity shifting rules on syntactic structures. However, sentiment accuracies for classifying a sentence as positive/negative/neutral has not exceeded 60$\%$. 

Early attempts used general subjectivity clues to generate training data from un-annotated text \cite{RSM:Ril2003}. Next, bag-of-words (BOW) classifiers were introduced that represent a document as a multi set of its words disregarding grammar and word order. These methods did not work well on short tweets. Co-occurrence matrices also were unable to capture difference in antonyms such as `good/bad' that have similar distributions. 
Subjectivity detection hence progressed from syntactic to semantic methods in \cite{RSM:Ril2003}, where the authors used extraction pattern to represent subjective expressions. For example, the pattern `hijacking' of $<x>$, looks for the noun `hijacking' and the object of the preposition $<x>$. Extracted features are used to train machine-learning classifiers such as SVM \cite{RSM:Wie2005} and ELM \cite{onesta}. Subjectivity detection is also useful for constructing and maintaining sentiment lexicons, as objective words or concepts need to be omitted from them \cite{camnt4}. 

Since, subjective sentences tend to be longer than neutral sentences, recursive neural networks were proposed where the sentiment class at each node in the parse tree was captured using matrix multiplication of parent nodes \cite{RSM:Nal2014,RSM:Xav2011}. However, the number of possible parent composition functions is exponential, hence in \cite{RSM:Ric2013} recursive neural tensor network was introduced that use a single tensor composition function to define multiple bilinear dependencies between words. 
In \cite{Maas:2011}, the authors used logistic regression predictor that defines a hyperplane in the word vector space where a word vectors positive sentiment probability depends on where it lies with respect to this hyperplane. However, it was found that while incorporating words that are more subjective can generally yield better results, the performance gain by employing extra neutral words is less significant \cite{ijcnlp2011}.
Another class of probabilistic models called Latent Dirichlet Allocation assumes each document is a mixture of latent topics. 
Lastly, sentence-level subjectivity detection was integrated into document-level sentiment detection using graphs where each node is a sentence. The contextual constraints between sentences in a graph led to significant improvement in polarity classification \cite{RSM:Pang2004}. 

Similarly, in \cite{RSM:Jun2006} the authors take advantage of the sequence encoding method for trees and treat them as sequence kernels for sentences.
Templates are not suitable for semantic role labeling, because relevant context might be very far away. Hence, deep neural networks have become popular to process text. In word2vec, for example, a word's meaning is simply a signal that helps to classify larger entities such as documents. Every word is mapped to a unique vector, represented by a column in a weight matrix. The concatenation or sum of the vectors is then used as features for prediction of the next word in a sentence \cite{RSM:Coll2011}.
Related words appear next to each other in a $d$ dimensional vector space. Vectorizing them allows us to measure their similarities and cluster them. For semantic role labeling, we need to know the relative position of verbs, hence the features can include prefix, suffix, distance from verbs in the sentence etc. However, each feature has a corresponding vector representation in $d$ dimensional space learned from the training data. 

Recently, convolutional neural network (CNN) is being used for subjectivity detection. In particular, \cite{RSM:Nal2013} used recurrent CNNs. These show high accuracy on certain datasets such as Twitter we are also concerned with a specific sentence within the context of the previous discussion, the order of the sentences preceding the one at hand results in a sequence of sentences also known as a time series of sentences \cite{RSM:Nal2013}. However, their model suffers from overfitting, hence in this work we consider deep convolutional neural networks, where temporal information is modeled via dynamic Gaussian Bayesian networks.

\section{Aspect-Based Sentiment Analysis}
\label{asoa}
Aspect extraction from opinions was first studied by \cite{Hu2004}.
They introduced the distinction between explicit and implicit aspects. 
However, the authors only dealt with explicit aspects and used a set of rules based on statistical observations. 
Hu and Liu's method was later improved by \cite{Popescu2005} and by \cite{Blair-Goldensohn2008}. 
\cite{Popescu2005} assumed the product class is known in advance. Their algorithm detects whether a noun or noun phrase is a product feature by computing the point-wise mutual information between the noun phrase and the product class.

\cite{scaffidi2007red} presented a method that uses language model to identify product features. They assumed that product features are more frequent in product reviews than in a general natural language text. However, their method seems to have low precision since retrieved aspects are affected by noise.
Some methods treated the aspect term extraction as sequence labeling and used CRF for that. Such methods have performed very well on the datasets even in cross-domain experiments \cite{jakob2010extracting,zhiqiang2014dlirec}.

Topic modeling has been widely used as a basis to perform extraction and grouping of aspects \cite{hu2014interactive,chen2014mining}. Two models were considered: pLSA \cite{hofmann1999probabilistic} and LDA \cite{blei2003latent}. Both models introduce a latent variable ``topic'' between the observable variables ``document'' and ``word'' to analyze the semantic topic distribution of documents. In topic models, each document is represented as a random mixture over latent topics, where each topic is characterized by a distribution over words. 

Such methods have been gaining popularity in social media analysis like emerging political topic detection in Twitter \cite{rilpol}.
The LDA model defines a Dirichlet probabilistic generative process for document-topic distribution; in each document, a latent aspect is chosen according to a multinomial distribution, controlled by a Dirichlet prior $\alpha$. Then, given an aspect, a word is extracted according to another multinomial distribution, controlled by another Dirichlet prior $\beta$.
Among existing works employing these models are the extraction of global aspects ( such as the brand of a product) and local aspects (such as the property of a product \cite{titov2008modeling}), the extraction of key phrases \cite{branavan2009learning}, the rating of multi-aspects \cite{wang2010latent}, and the summarization of aspects and sentiments \cite{lu2009rated}. 
\cite{zhao2010jointly} employed the maximum entropy method to train a switch variable based on POS tags of words and used it to separate aspect and sentiment words. 

\cite{mcauliffe2008supervised} added user feedback to LDA as a response-variable related to each document. \cite{lu2008opinion} proposed a semi-supervised model. DF-LDA \cite{andrzejewski2009incorporating} also represents a semi-supervised model, which allows the user to set must-link and cannot-link constraints. A must-link constraint means that two terms must be in the same topic, while a cannot-link constraint means that two terms cannot be in the same topic. \cite{porlda} integrated commonsense in the
calculation of word distributions in the LDA algorithm, thus
enabling the shift from syntax to semantics in aspect-based
sentiment analysis. 
\cite{wang2014product} proposed two semi-supervised models for product aspect extraction based on the use of seeding aspects. In the category of supervised methods, \cite{jagarlamudi2012incorporating} employed seed words to guide topic models to learn topics of specific interest to a user, while \cite{wang2010latent} and \cite{mukherjee2012aspect} employed seeding words to extract related product aspects from product reviews.
On the other hand, recent approaches using deep CNNs \cite{collobert2011natural,poria-cambria-gelbukh:2015:EMNLP} showed significant performance improvement over the state-of-the-art methods on a range of NLP tasks. \cite{collobert2011natural} fed word embeddings to a CNN to solve standard NLP problems such as named entity recognition (NER), part-of-speech (POS) tagging and semantic role labeling.

\section{Preliminaries}
\label{prep}
In this section, we briefly review the theoretical concepts necessary to comprehend the present work. We begin with a description of maximum likelihood estimation of edges in dynamic Gaussian Bayesian networks where each node is a word in a sentence. Next, we show that weights in the CNN can be learned by minimizing a global error function that corresponds to an exponential distribution over a linear combination of input sequence of word features.

\textbf{Notations :} Consider a Gaussian network (GN) with time delays which comprises a set of $\mathrm{N}$ nodes and observations gathered over $\mathrm{T}$ instances for all the nodes. Nodes can take real values from a multivariate distribution determined by the parent set. Let the dataset of samples be $X={\{x_{i}(t)\}}_{\mathrm{N}\times \mathrm{T}}$, where $x_{i}(t)$ represents the sample value of the $i^{\textrm{ th}}$ random variable in instance $t$. Lastly, let $\boldsymbol{a}_{i}$ be the set of parent variables regulating variable $i$.

\subsection{Gaussian Bayesian Networks}

In tasks where one is concerned with a specific sentence within the context of the previous discourse, capturing the order of the sequences preceding the one at hand may be particularly crucial. 

We take as given a sequence of sentences $s(1),s(2),\ldots,s(T)$, each in turn being a sequence of words so that $s(t)=(x_{1}(t),x_{2}(t),\ldots,x_{\mathrm{L}}(t))$, where $\mathrm{L}$ is the length of sentence $s(t)$. Thus, the probability of a word $p(x_{i}(t))$ follows the distribution :
\begin{eqnarray}
p(x_{i}(t))=P(x_{i}(t)|(x_{1}(t),x_{2}(t),\\
\nonumber
\ldots,x_{i-1}(t)),(s(1),s(2),\ldots,s(t-1))
\label{eq1}
\end{eqnarray} 

A Bayesian network is a graphical model that represents a joint multivariate probability distribution for a set of random variables \cite{RSM:Pri2009}. It is a directed acyclic graph $S$ with a set of parameters $\boldsymbol{\theta}$ that represents the strengths of connections by conditional probabilities.
\begin{figure}[ht]\begin{center}
\begin{tabular}{c}
\includegraphics[height=6cm]{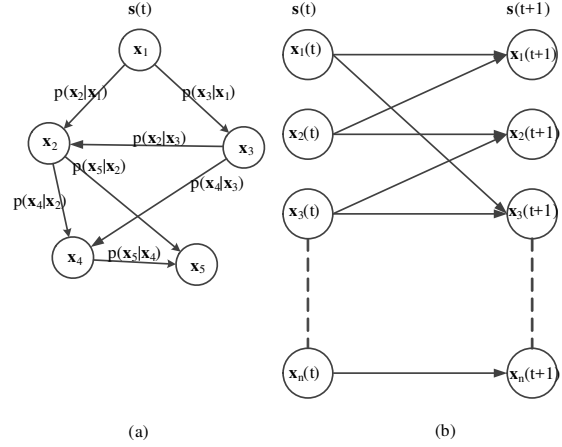}
 \end{tabular}
\caption{State space of different Bayesian models}
  \label{fig:bn1}
 \end{center}
 \end{figure}

The BN decomposes the likelihood of node expressions into a product of
conditional probabilities by assuming independence of
non-descendant nodes, given their parents.
\begin{eqnarray}
p(X|S,\boldsymbol{\theta})={\prod}_{i=1}^{\mathrm{N}}p({\boldsymbol{\mathrm{x}}}_{i}|
{\boldsymbol{a}}_{i}, {\theta}_{i,{\boldsymbol{a}}_{i}}),
\label{eq:bn}
\end{eqnarray}
\noindent
where $p(\boldsymbol{\mathrm{x}}_{i}|\boldsymbol{a}_{i},\theta_{i,\boldsymbol{a}_{i}})$ denotes the
conditional probability of node expression $\boldsymbol{\mathrm{x}}_{i}$ given its
parent node expressions $\boldsymbol{a}_{i}$, and $\theta_{i,\boldsymbol{a}_{i}}$ denotes the maximum likelihood(ML) estimate of the conditional probabilities. 

Figure ~\ref{fig:bn1} (a) illustrates the state space of a Gaussian Bayesian network (GBN) at time instant $t$ where each node $x_{i}(t)$ is a word in the sentence $s(t)$. The connections represent causal dependencies over one or more time instants. The observed state vector of variable $i$ is denoted as $\boldsymbol{\mathrm{x}}_i$ and the conditional probability of variable $i$ given variable $j$ is $p(\boldsymbol{\mathrm{x}}_{i}|\boldsymbol{\mathrm{x}}_{j})$.
The optimal Gaussian network $S^{*}$ is obtained by maximizing the posterior probability of $S$ given the data $X$. From Bayes theorem, the optimal Gaussian network $S^{*}$ is given by:
\begin{eqnarray}
S^{*}=\mathrm{arg} \max_{S} p(S|X) = \mathrm{arg} \max_{S} p(S)p(X|S), \label{eqn:bayes}
\end{eqnarray}
\noindent
where $p(S)$ is the probability of the Gaussian network and
$p(X|S)$ is the likelihood of the expression data given the Gaussian network.

Given the set of conditional distributions with parameters $\boldsymbol{\theta}
= \{ \theta_{i,\boldsymbol{a}_{i}}\}_{i=1}^{\mathrm{N}}$, the likelihood of the data is given by
\begin{eqnarray}
p(X|S)=\int p({X}|S,\boldsymbol{\theta})p(\boldsymbol{\theta}|S)d\theta, \label{eq:int}
\end{eqnarray}
\noindent
To find the likelihood in (\ref{eq:int}), and to obtain the
optimal Gaussian network as in (\ref{eqn:bayes}), Gaussian BN assumes that the nodes are multivariate Gaussian. That is, expression of node $i$ can be described with mean $\mu_{i}$ and
covariance matrix $\Sigma_{i}$ of size $\mathrm{N} \times \mathrm{N}$. The joint probability of the network can be
the product of a set of conditional probability distributions given by:
\begin{eqnarray}
p(\boldsymbol{\mathrm{x}}_{i}|\boldsymbol{a}_{i})={\theta}_{i,\boldsymbol{a}_{i}}\sim\cal{N}\Bigg( \textrm{$\mu_{i}+\sum_{j\in \boldsymbol{a}_{i}} (\boldsymbol{\mathrm{x}}_{j}-\mu_{j})\beta$, $\Sigma_{i}^{'}$} \Bigg),
\label{gau}
\end{eqnarray}
\noindent
where $\Sigma_{i}^{'}=\Sigma_{i}-\Sigma_{i,\boldsymbol{a}_{i}}\Sigma_{\boldsymbol{a}_{i}}^{-1}\Sigma_{i,\boldsymbol{a}_{i}}^{T}$ and $\beta$ denotes the regression coefficient matrix, $\Sigma_{i}^{'}$ is the conditional variance of $\boldsymbol{\mathrm{x}}_{i}$ given its
parent set $\boldsymbol{a}_{i}$, $\Sigma_{i,\boldsymbol{a}_{i}}$ is the covariance
between observations of $\boldsymbol{\mathrm{x}}_{i}$ and the variables in $\boldsymbol{a}_{i}$, and
$\Sigma_{\boldsymbol{a}_{i}}$ is the covariance matrix of $\boldsymbol{a}_{i}$.
The acyclic condition of BN does not allow feedback among nodes, and feedback is an essential characteristic of real world GN. 

Therefore, dynamic Bayesian networks have recently become popular in building
GN with time delays mainly due to their ability to model causal interactions as well as feedback regulations
\cite{RSM:Fri98}. A first-order dynamic BN is defined by a transition network of interactions between a pair of Gaussian networks connecting nodes at time instants $\tau$ and $\tau+1$. In time instant $\tau+1$, the parents of nodes are those specified in the time instant $\tau$. 
Similarly, the Gaussian network of a $\mathrm{R}$-order dynamic system is represented by a Gaussian network comprising $(\mathrm{R}+1)$ consecutive time points and $\mathrm{N}$ nodes, or a graph of $(\mathrm{R}+1)\times \mathrm{N}$ nodes. In practice, the sentence data is transformed to a BOW model where each sentence is a vector of frequencies for each word in the vocabulary. 
Figure ~\ref{fig:bn1} (b) illustrates the state space of a first-order Dynamic GBN models transition networks among words in sentences $s(t)$ and $s(t+1)$ in consecutive time points, the lines correspond to first-order edges among the words learned using BOW. 

Hence, a sequence of sentences results in a time series of word frequencies. It can be seen that such a discourse model produces compelling discourse vector representations that are sensitive to the structure of the discourse and promise to capture subtle aspects of discourse comprehension, especially when coupled to further semantic data and unsupervised pre-training. 

\subsection{Convolutional Neural Networks}
\label{sec:background-on-deep-CNN}
The idea behind convolution is to take the dot product of a vector of $k$ weights $w_{k}$ also known as kernel vector with each $k$-gram in the sentence $s(t)$ to obtain another sequence of features $c(t)=(c_{1}(t),c_{2}(t),\ldots,c_{\mathrm{L}}(t))$. 
\begin{eqnarray}
{c}_{j}={{w}_{k}}^{T}.{\boldsymbol{\mathrm{x}}_{i:i+k-1}}
\label{eqn:conv1}
\end{eqnarray}
We then apply a max pooling operation over the feature map and take the maximum value $\hat{c}(t)=\mathrm{max}\{\boldsymbol{\mathrm{c}}(t)\}$ as the feature corresponding to this particular kernel vector. Similarly, varying kernel vectors and window sizes are used to obtain multiple features \cite{RSM:Nal2014}. 

For each word $x_{i}(t)$ in the vocabulary, an $d$ dimensional vector representation is given in a look up table that is learned from the data \cite{RSM:Coll2011}. The vector representation of a sentence is hence a concatenation of vectors for individual words. Similarly, we can have look up tables for other features. One might want to provide features other than words if these features are suspected to be helpful. Now, the convolution kernels are applied to word vectors instead of individual words. 

We use these features to train higher layers of the CNN that can represent bigger groups of words in sentences. We denote the feature learned at hidden neuron $h$ in layer $l$ as $F^{l}_{h}$. Multiple features may be learned in parallel in the same CNN layer. The features learned in each layer are used to train the next layer 
\begin{eqnarray}
F^{l}={\sum}_{h=1}^{n_{h}}w_{k}^{h}*F^{l-1}
\end{eqnarray}
where * indicates convolution and $w_{k}$ is a weight kernel for hidden neuron $h$ and $n_{h}$ is the total number of hidden neurons.
Training a CNN becomes difficult as the number of layers increases, as the Hessian matrix of second-order derivatives often does not exist. Recently, deep learning has been used to improve the scalability of a model that has inherent parallel computation. This is because hierarchies of modules can provide a compact representation in the form of input-output pairs. 
Each layer tries to minimize the error between the original state of the input nodes and the state of the input nodes predicted by the hidden neurons. 

This results in a downward coupling between modules. The more abstract representation at the output of a higher layer module is combined with the less abstract representation at the internal nodes from the module in the layer below. In the next section, we describe deep CNN that can have arbitrary number of layers.

\subsection{Convolution Deep Belief Network}
A deep belief network (DBN) is a type of deep neural network that can be viewed as a composite of simple, unsupervised models such as restricted Boltzmann machines (RBMs) where each RBMs hidden layer serves as the visible layer for the next RBM \cite{chalea}. RBM is a bipartite graph comprising two layers of neurons: a visible and a hidden layer; it is restricted such that the connections among neurons in the same layer are not allowed. 
To compute the weights $W$ of an RBM, we assume that the probability distribution over the input vector $\boldsymbol{\mathrm{x}}$ is given as:
\begin{eqnarray}
p(\boldsymbol{\mathrm{x}}|W)=\frac{1}{Z(W)}{\exp}^{-\mathrm{E}(\boldsymbol{\mathrm{x}};W)}
\end{eqnarray}
\noindent

where ${Z(W)}={\sum}_{\boldsymbol{\mathrm{x}}}{\exp}^{-\mathrm{E}}(\boldsymbol{\mathrm{x}};W)$ is a normalisation constant. Computing the maximum likelihood is difficult as it involves solving the normalisation constant, which is a sum of an exponential number of terms. The standard approach is to approximate the average over the distribution with an average over a sample from $p(\boldsymbol{\mathrm{x}}|W)$, obtained by Markov chain Monte Carlo until convergence. 

To train such a multi-layer system, we must compute the gradient of the total energy function $\mathrm{E}$ with respect to weights in all the layers. To learn these weights and maximize the global energy function, the approximate maximum likelihood contrastive divergence (CD) approach can be used. This method employs each training sample to initialize the visible layer. Next, it uses the Gibbs sampling algorithm to update the hidden layer and then reconstruct the visible layer consecutively, until convergence \cite{RSMHin2002}. As an example, here we use a logistic regression model to learn the binary hidden neurons and each visible unit is assumed to be a sample from a normal distribution \cite{RSM:Gra2006}. 

The continuous state $\hat{h}_{j}$ of the hidden neuron $j$, with bias $b_{j}$, is a weighted sum over all continuous visible nodes $\boldsymbol{v}$ and is given by:
\begin{eqnarray}
\hat{h}_{j}=b_{j}+\sum_{i}v_{i}w_{ij}, \label{h2a}
\end{eqnarray}
\noindent

where $w_{ij}$ is the connection weight to hidden neuron $j$ from visible node $v_{i}$. The binary state $h_{j}$ of the hidden neuron can be defined by a sigmoid activation function:
\begin{eqnarray}
h_{j}=\frac{1}{1+e^{-\hat{h}_{j}}}.
\end{eqnarray}
\noindent

Similarly, in the next iteration, the binary state of each visible node is reconstructed and labeled as $\boldsymbol{v}_{recon}$. Here, we determine the value to the visible node $i$, with bias $c_{i}$, as a random sample from the normal distribution where the mean is a weighted sum over all binary hidden neurons and is given by:
\begin{eqnarray}
\hat{v}_{i}=c_{i}+\sum_{j}h_{i}w_{ij}, \label{h2}
\end{eqnarray}
\noindent

where $w_{ij}$ is the connection weight to hidden neuron $j$ from visible node $v_{i}$. The continuous state $v_{i}$ is a random sample from $\mathcal{N}(\hat{v}_{i},\sigma)$, where $\sigma$ is the variance of all visible nodes. Lastly, the weights are updated as the difference between the original and reconstructed visible layer using:
\begin{eqnarray}
\triangle
w_{ij}=\alpha(<v_{i}h_{j}>_{data}-<v_{i}h_{j}>_{recon}),
\label{cd}
\end{eqnarray}
\noindent
where $\alpha$ is the learning rate and $<v_{i}h_{j}>$ is the
expected frequency with which visible unit $i$ and hidden unit $j$
are active together when the visible vectors are sampled from the
training set and the hidden units are determined by (\ref{h2a}). Finally, the energy of a DNN can be determined in the final layer using $\mathrm{E}=-\sum_{i,j}v_{i}h_{j}w_{ij}$.

To extend the deep belief networks to convolution deep belief network (CDBN) we simply partition the hidden layer into $\mathrm{Z}$ groups. Each of the $\mathrm{Z}$ groups is associated with a $k \times d$ filter where $k$ is the width of the kernel and $d$ is the number of dimensions in the word vector. Let us assume that the input layer has dimension $\mathrm{L}\times d$ where $\mathrm{L}$ is the length of the sentence. Then the convolution operation given by (\ref{eqn:conv1}) will result in a hidden layer of $\mathrm{Z}$ groups each of dimension ${(\mathrm{L}-k+1)}\times{(d-d+1)}$. These learned kernel weights are shared among all hidden units in a particular group. The energy function is now a sum over the energy of individual blocks given by:
\begin{eqnarray}
\mathrm{E}=-\sum_{z=1}^{\mathrm{Z}}\sum_{i,j}^{\mathrm{L}-k+1,1}\sum_{r,s}^{k,d}v_{i+r-1,j+s-1}h_{ij}^{z}w_{rs}^{k}
\end{eqnarray}

The CNN sentence model preserve the order of words by adopting convolution kernels of gradually increasing sizes that span an increasing number of words and ultimately the entire sentence \cite{RSM:Nal2013}. However, several word dependencies may occur across sentences hence, in this work we propose a Bayesian CNN model that uses dynamic Bayesian networks to model a sequence of sentences.

\section{Deep Learning Algorithm}
\label{algo}
\subsection{ Subjectivity Detection}
In this work, we integrate a higher-order GBN for sentences into the first layer of the CNN. The GBN layer of connections $\beta$ is learned using maximum likelihood approach on the BOW model of the training data. The input sequence of sentences $s(t:t-2)$ are parsed through this layer prior to training the CNN. Only sentences or groups of sentences containing high ML motifs are then used to train the CNN.
Hence, motifs are convolved with the input sentences to generate a new set of sentences for pre-training.
\begin{eqnarray}
F^{0}={\sum}_{h=1}^{M}{\beta}^{h}*\boldsymbol{s}
\label{eqn:bcnn}
\end{eqnarray}
where $M$ is the number of high ML motifs and $\boldsymbol{s}$ is the training set of sentences in a particular class. 

\begin{figure*}[ht]\begin{center}
\begin{tabular}{c}
\includegraphics[height=7.4cm]{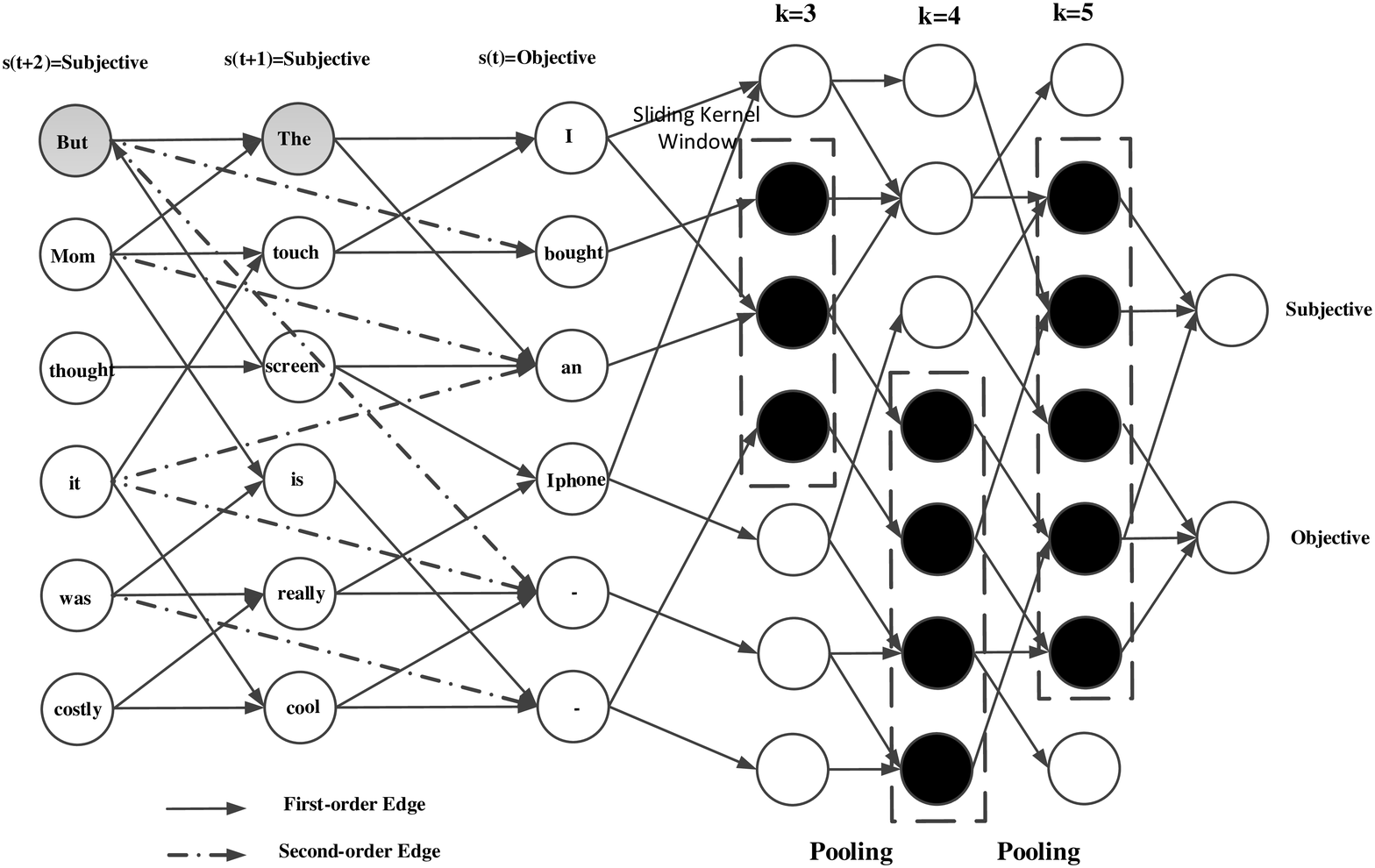}
 \end{tabular}
\caption{State space of Bayesian CNN where the input layer is pre-trained using a dynamic GBN}
  \label{fig:bn2}
 \end{center}
 \end{figure*}

Fig.~\ref{fig:bn2} illustrates the state space of Bayesian CNN where the input layer is pre-trained using a dynamic GBN with up-to two time point delays shown for three sentences in a review on iPhone. The dashed lines correspond to second-order edges among the words learned using BOW. Each hidden layer does convolution followed by pooling across the length of the sentence. To preserve the order of words we adopt kernels of increasing sizes.

Since, the number of possible words in the vocabulary is very large, we consider only the top subjectivity clue words to learn the GBN layer. Lastly, In-order to preserve the context of words in conceptual phrases such as `touchscreen'; we consider additional nodes in the Bayesian network for phrases with subjectivity clues. Further, the word embeddings in the CNN are initialized using the log-bilinear language model (LBL) where the $d$ dimensional vector representation of each word $x_{i}(t)$ in (\ref{eq1}) is given by :
\begin{eqnarray}
{x}_{i}(t)={\sum}_{k=1}^{i-1}{C}_{k}{x}_{k}(t)
\end{eqnarray}
\noindent
where $C_{k}$ are the $d\times d$ co-occurrence or context matrices computed from the data.

The time series of sentences is used to generate a sub-set of sentences containing high ML motifs using (\ref{eqn:bcnn}). The frequency of a sentence in the new dataset will also correspond to the corresponding number of high ML motifs in the sentence. In this way, we are able to increase the weights of the corresponding causal features among words and concepts extracted using Gaussian Bayesian networks.

The new set of sentences is used to pre-train the deep neural network prior to training with the complete dataset.
Each sentence can be divided into chunks or phrases using POS taggers. The phrases have hierarchical structures and combine in distinct ways to form sentences. The $k$-gram kernels learned in the first layer hence correspond to a chunk in the sentence. 
\subsection{Aspect Extraction}
\label{training}
In order to train the CNN for aspect extraction, instead, we used a special training algorithm suitable for sequential data, proposed by \cite{collobert2011natural}. We will summarize it here, mainly following \cite{nlpnet}.
The algorithm trains the neural network by back-propagation in order to maximize the likelihood over training sentences. 
Consider the network parameter $\theta$. We say that $h_y$ is the output score for the likelihood of an input $x$ to have the tag $y$. Then, the probability to assign the label $y$ to $x$ is calculated as
\begin{align}
	p(y|x,\theta) = \frac{\exp{(h_y)}}{\sum_j\exp{(h_j)}}.
\end{align}
Define the logadd operation as
\begin{align}
	\logadd_ih_{i} = \log{\sum_i\exp{h_i}},\label{eq:log}
\end{align}
then for a training example, the log-likelihood becomes
\begin{align}
	\log p(y|x,\theta) = h_y - \logadd_ih_{i}.
\end{align}
In aspect term extraction, the terms can be organized as chunks and are also often surrounded by opinion terms. Hence, it is important to consider sentence structure on a whole in order to obtain additional clues. Let it be given that there are $T$ tokens in a sentence and $y$ is the tag sequence while $h_{t,i}$ is the network score for the $t$-th tag having $i$-th tag. We introduce $A_{i,j}$ transition score from moving tag $i$ to tag $j$. Then, the score tag for the sentence $s$ to have the tag path $y$ is defined by:
\begin{align}
	s(x,y,\theta) = \sum_{t=1}^{T}(h_{t,y_{t}} + A_{y_{t-1},y_{t}}). \label{minimize}
\end{align}
This formula represents the tag path probability over all possible paths. Now, from~\eqref{eq:log} we can write the log-likelihood 
\begin{align}
	\log p(y|x,\theta) = s(x,y,\theta) - \logadd_{\forall j}s(x,j,\theta). \label{log}
\end{align}
The number of tag paths has exponential growth. However, using dynamic programming techniques, one can compute in polynomial time the score for all paths that end in a given tag \cite{collobert2011natural}. 
Let $y_{t}^{k}$ denote all paths that end with the tag $k$ at the token $t$. Then, using recursion, we obtain 
\begin{align}
	\delta_{t}(k) = \logadd_{\forall y_{t}^{k}}s(x,y_{t}^{k},\theta) = h_{t,k} + \logadd_j\delta_{t-1}(j) + A_{j,k}.
\end{align}
For the sake of brevity, we shall not delve into details of the recursive procedure, which can be found in \cite{collobert2011natural}. 
The next equation gives the log-add for all the paths to the token $T$:
\begin{align}
	\logadd_{\forall y}s(x,y,\theta) = \logadd_i\delta_T(i).
\end{align} 

Using these equations, we can maximize the likelihood of ~\eqref{log} over all training pairs. For inference, we need to find the best tag path using the Viterbi algorithm; e.g., we need to find the best tag path that minimizes the sentence score~\eqref{minimize}.

The features of an aspect term depend on its surrounding words. Thus, we used a window of 5 words around each word in a sentence, i.e., $\pm2$ words. We formed the local features of that window and considered them to be features of the middle word.
Then, the feature vector was fed to a CNN. 

The network contained one input layer, two convolution layers, two max-pool layers, and a fully connected layer with softmax output. The first convolution layer consisted of 100 feature maps with filter size 2. The second convolution layer had 50 feature maps with filter size 3. The stride in each convolution layer is 1 as we wanted to tag each word. A max-pooling layer followed each convolution layer. The pool size we use in the max-pool layers was 2. We used regularization with dropout on the penultimate layer with a constraint on L2-norms of the weight vectors, with 30 epochs.
The output of each convolution layer was computed using a non-linear function; in our case we used $\tanh$.

As features, we used word embeddings trained on two different corpora. We also used some additional features and rules to boost the accuracy; see Section~\ref{sec:features-and-rules}. The CNN produces local features around each word in a sentence and then combines these features into a global feature vector. 
Since the kernel size for the two convolution layers was different, the dimensionality $L_x \times L_y$ mentioned in Section~\ref{sec:background-on-deep-CNN} was $3\times300$ and $2\times300$, respectively. The input layer was $65\times300$, where 65 was the maximum number of words in a sentence, and 300 the dimensionality of the word embeddings used, per each word.

The process was performed for each word in a sentence. Unlike traditional max-likelihood leaning scheme, we trained the system using propagation after convolving all tokens in the sentence. Namely, we stored the weights, biases, and features for each token after convolution and only back-propagated the error in order to correct them once all tokens were processed using the training scheme as explained in Section~\ref{training}. 

If a training instance $s$ had $n$ words, then we represented the input vector for that instance as $s_{1:n} = s_{1}\bigoplus s_{2}\bigoplus ... \bigoplus s_{n}$. Here, $s_i \in \Re^k$ is a $k$-dimensional feature vector for the word~$s_i$. We found that this network architecture produced good results on both of our benchmark datasets. Adding extra layers or changing the pooling size and window size did not contribute to the accuracy much, and instead, only served to increase computational cost. 

\section{Evaluation}
\label{eval}

\subsection{Subjectivity Detection}
\subsubsection{Datasets Used}
We use the MPQA corpus \cite{RSM:Wie2005}, a collection of 535 English news articles from a variety of sources manually annotated with subjectivity flag. From the total of 9,700 sentences in this corpus, 55$\%$ of the sentences are labeled as subjective while the rest are objective. We also compare with the Movie Review (MR) benchmark dataset \cite{RSM:Pang2004}, that contains 5000 subjective movie review snippets from Rotten Tomatoes website and another 5000 objective sentences from plot summaries available from the Internet Movies Database. All sentences are at least ten words 
long and drawn from reviews or plot summaries of movies released post 2001.

The data pre-processing included removing top 50 stop words and punctuation marks from the sentences. Next, we used a POS tagger to determine the part-of-speech for each word in a sentence. Subjectivity clues dataset \cite{RSM:Ril2003} contains a list of over 8,000 clues identified manually as well as automatically using both annotated and unannotated data. Each clue is a word and the corresponding part of speech. 

The frequency of each clue was computed in both subjective and objective sentences of the MPQA corpus. Here we consider the top 50 clue words with highest frequency of occurrence in the subjective sentences. We also extracted 25 top concepts containing the top clue words using the method described in \cite{porsent}.
The CNN is collectively pre-trained with both subjective and objective sentences that contain high ML word and concept motifs. The word vectors are initialized using the LBL model and a context window of size 5 and 30 features. Each sentence is wrapped to a window of 50 words to reduce the number of parameters and hence the over-fitting of the model. A CNN with three hidden layers of 100 neurons and kernels of size $\{3,4,5\}$ is used. The output layer corresponds to two neurons for each class of sentiments. 
\subsubsection{Experimental Results}
We used 10 fold cross validation to determine the accuracy of classifying new sentences using the trained CNN classifier. A comparison is done with classifying the time series data using baseline classifiers such as Naive Bayes SVM (NBSVM) \cite{RSM:Wan2013}, Multichannel CNN (CNN-MC) \cite{RSM:Yoon2014}, Subjectivity Word Sense Disambiguation (SWSD) \cite{RSM:Ort2013} and Unsupervised-WSD (UWSD) \cite{RSM:Akk2009}. Table \ref{deep1} shows that BCDBN outperforms previous methods by $5-10\%$ in accuracy on both datasets. Almost $10\%$ improvement is observed over NBSVM on the movie review dataset. In addition, we only consider word vectors of 30 features instead of the 300 features used by CNN-MC and hence are 10 times faster. 
\begin{table}[h]
	\centering
	{\scriptsize
		\caption{ 
			F-measure by different models for classifying sentences in a document as Subjective and Objective in MPQA and MR dataset. } 
		\begin{tabular}{c|c|c|c|c|c}
			\hline
			Dataset & NBSVM & CNN-MC & SWSD & UWSD & BCDBN \\
			\hline
			MPQA & 86.3 & 89.4 & 80.35 & 60 & \bf{93.2} \\
			MR & 93.2 & 93.6 & - & 55 & \bf{96.4} \\
			\hline
		\end{tabular}
	}
	\label{deep1}
\end{table}
\subsection{Aspect Extraction}

\subsubsection{Datasets Used}
In this subsection, we present the data used in our experiments.

\subsubsection{Google Embeddings}
\cite{mikolov2013linguistic} presented two different neural network models for creating word embeddings. The models were log-linear in nature, trained on large corpora. One of them is a bag-of-words based model called CBOW; it uses word context in order to obtain the embeddings. The other one is called skip-gram model; it predicts the word embeddings of surrounding words given the current word. Those authors made a dataset called word2vec publicly available. These 300-dimensional vectors were trained  on  a 100-billion-word corpus from Google News using the CBOW architecture.

\subsubsection{Our Amazon Embeddings}
We trained the CBOW architecture proposed by \cite{mikolov2013linguistic} on a large Amazon product review dataset developed by \cite{mcauley2013hidden}. This dataset consists of 34,686,770 reviews (4.7 billion words) of 2,441,053 Amazon products from June 1995 to March 2013. We kept the word embeddings 300-dimensional (http://sentic.net/AmazonWE.zip).
Due to the nature of the text used to train this model, this includes opinionated/affective information, which is not present in ordinary texts such as the Google News corpus.

\subsubsection{Evaluation Corpora}\label{corpus}
For training and evaluation of the proposed approach, we used two corpora:
\begin{itemize}
\item Aspect-based sentiment analysis dataset developed by \cite{qiu2011opinion}; and
\item SemEval 2014 dataset. The dataset consists of training and test sets from two domains, Laptop and Restaurant; see Table~\ref{semeval}.
\end{itemize}

The annotations in both corpora were encoded according to IOB2, a widely used coding scheme for representing sequences. In this encoding, the first word of each chunk starts with a ``B-Type'' tag, ``I-Type'' is the continuation of the chunk and ``O'' is used to tag a word which is out of the chunk. In our case, we are interested to determine whether a word or chunk is an aspect, so we only have ``B--A'', ``I--A'' and ``O'' tags for the words.\\ 

Here is an example of IOB2~tags:
\begin{quote}
\raggedright
\emph{also/O excellent/O operating/B-A system/I-A~,/O size/B-A and/O weight/B-A for/O optimal/O mobility/B-A excellent/O durability/B-A of/O the/O battery/B-A the/O functions/O provided/O by/O the/O trackpad/B-A is/O unmatched/O by/O any/O other/O brand/O}
\end{quote}

\subsubsection{Features and Rules Used}\label{sec:features-and-rules}

In this section, we present the features, the representation of the text, and linguistic rules used in our experiments.

We used the following the features:
\begin{itemize}
\item{\bf Word Embeddings} ~ We used the word embeddings described earlier as features for the network. This way, each word was encoded as 300-dimensional vector, which was fed to the network.
\item{\bf Part of speech tags} ~ Most of the aspect terms are either nouns or noun chunk. This justifies the importance of POS features. We used the POS tag of the word as its additional feature. We used 6 basic parts of
speech (noun, verb, adjective, adverb, preposition,
conjunction) encoded as a 6-
dimensional binary vector. We used Stanford
Tagger as a POS tagger.

These two features vectors were concatenated and fed to CNN. 

So, for each word the final feature vector is 306 dimensional.
\end{itemize}

\begin{table}[t!]
\centering
\begin{tabular}{l crr}
\hline\bigstrut
Domain & Training & Test & Total\\
\hline\bigstrut[t]
Laptop & 3041 & 800 & 3841\\
Restaurant & 3045 & 800 & 3845
\bigstrut[b]\\\hline%\hline
\bigstrut
Total & 6086 & 1600 & 7686 \\
\hline%\hline
\end{tabular}
\caption{SemEval Data used for Evaluation}
\label{semeval}
\end{table}

In some of our experiments, we used a set of linguistic patterns (LPs) derived from sentic patterns (LP) \cite{porsent}, a linguistic framework based on SenticNet \cite{camnt4}. SenticNet is a concept-level knowledge base for sentiment analysis built by means of sentic computing \cite{camsen}, a multi-disciplinary approach to natural language processing and understanding at the crossroads between affective computing, information extraction, and commonsense reasoning, which exploits both computer and human sciences to better interpret and process social information on the Web. In particular, we used the following linguistic rules:

\begin{description}
\item[Rule 1] Let a noun \emph{h} be a subject of a word \emph{t}, which has an adverbial or adjective modifier present in a large sentiment lexicon, SenticNet. Then mark \emph{h} as an aspect.
\item[Rule 2] Except when the sentence has an auxiliary verb, such as \textit{is}, \textit{was}, \textit{would}, \textit{should}, \textit{could}, etc., we apply:
\begin{description}
\item[Rule 2.1] If the verb \emph{t} is modified by an adjective or adverb or is in adverbial clause modifier relation with another token, then mark \emph{h} as an aspect.
E.g., in ``The battery lasts little'', 

\emph{battery} is the subject of \emph{lasts}, which is modified by an adjective modifier \emph{little}, so \emph{battery} is marked as an aspect.

\item[Rule 2.2] If \emph{t} has a direct object, a noun \emph{n}, not found in SenticNet, then mark \emph{n} an aspect, as, e.g., in ``I like the lens of this camera''.

\end{description}
\item[Rule 3] If a noun \emph{h} is a complement of a couplar verb, then mark \emph{h} as an explicit aspect.
E.g., in ``The camera is nice'',
% \ref{ex100} 
\emph{camera} is marked as an aspect.

\item[Rule 4] If a term marked as an aspect by the CNN or the other rules is in a noun-noun compound relationship with another word, then instead form one aspect term composed of both of them. E.g., if in ``battery life'', ``battery'' or ``life'' is marked as an aspect, then the whole expression is marked as an aspect.
 
\item[Rule 5] The above rules 1--4 improve recall by discovering more aspect terms. However, to improve precision, we apply some heuristics: e.g., we remove stop-words such as \emph{of}, \emph{the}, \emph{a}, etc., even if they were marked as aspect terms by the CNN or the other rules.

\end{description} 

We used the Stanford parser to determine syntactic relations in the sentences.

We combined LPs with the CNN as follows: both LPs and CNN-based classifier are run on the text; then all terms marked by any of the two classifiers are reported as aspect terms, except for those unmarked by the last rule.

\subsubsection{Experimental Results}\label{sec:results}
Table \ref{semeval1} shows the accuracy of our aspect term extraction framework in laptop and restaurant domains. The framework gave better accuracy on restaurant domain reviews, because of the lower variety of aspect available terms than in laptop domain. However, in both cases recall was lower than precision. 

Table \ref{semeval1} shows improvement in terms of both precision and recall when the POS feature is used.
Pre-trained word embeddings performed better than randomized features (each word's vector initialized randomly); see Table \ref{semeval4}. Amazon embeddings performed better than Google word2vec embeddings. This supports our claim that the former contains opinion-specific information which helped it to outperform the accuracy of Google embeddings trained on more formal text---the Google news corpus. 
Because of this, in the sequel we only show the performance using Amazon embeddings, which we denote simply as WE (word embeddings).

\makefigure{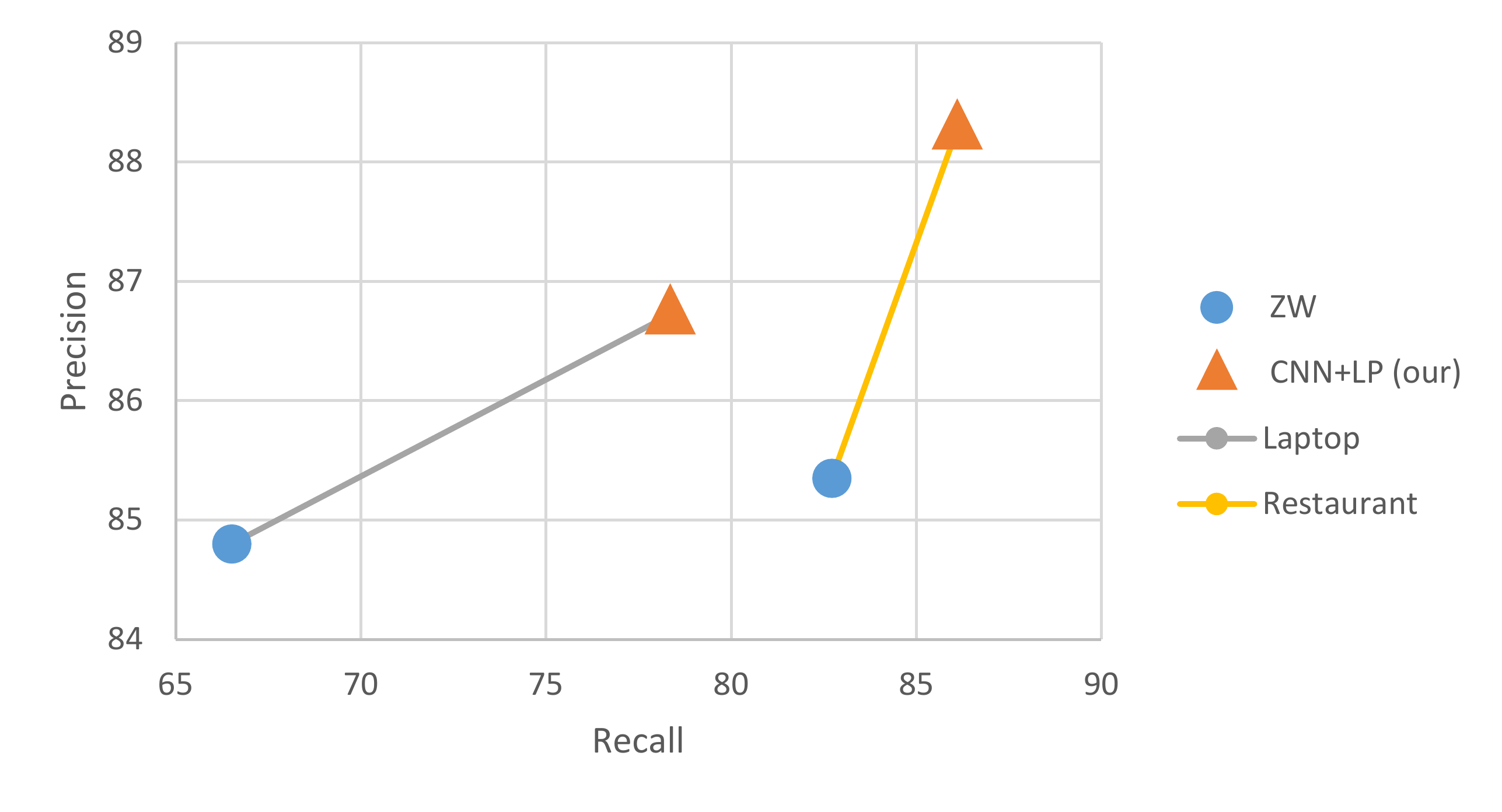}{Comparison of the performance with the state of the art.}

\begin{table}[h]
\centering

\begin{tabular}{llc}
\hline\bigstrut
Domain & Feature & F-Score\\
\hline\bigstrut[t]
Laptop & Random & 71.21\%\\
Laptop & Google Embeddings & 77.32\%\\
Laptop & Amazon Embeddings &  \bf 80.68\%\\
Restaurant & Random & 77.05\%\\
Restaurant & Google Embeddings & 83.50\%\\
Restaurant & Amazon Embeddings & \bf 85.70\%
\bigstrut[b]\\\hline%\hline
\end{tabular}
\caption{Random features vs. Google Embeddings vs. Amazon Embeddings on the SemEval 2014 dataset}
\label{semeval4}
\end{table}

\begin{table}[h]
\centering
\begin{tabular}{llccc}
\hline\bigstrut
Domain & Features & Recall & Precision & F-Score\\
\hline\bigstrut[t]
Laptop & WE & 75.20\% & 86.05\% & 80.68\%\\
Laptop & WE+POS & \bf 76.31\% & \bf 86.46\%  & \bf 81.06\%\\
Restaurant & WE & 84.11\% & 87.35\% & 85.70\%\\
Restaurant & WE+POS & \bf 85.01\% & \bf 87.42\% & \bf 86.20\%
\bigstrut[b]\\\hline%\hline
\end{tabular}
\caption{Feature analysis for the CNN classifier}
\label{semeval1}
\end{table}

In both domains, CNN suffered from low recall, i.e., it missed some valid aspect terms. Linguistic analysis of the syntactic structure of the sentences substantially helped to overcome some drawbacks of machine learning-based analysis. Our experiments showed good improvement in both precision and recall when LPs were used together with CNN; see Table~\ref{semeval2}.

\begin{table}[h]
\centering
\begin{tabular}{llccc}
\hline\bigstrut
Domain & Classifiers & Recall & Precision & F-Score\\
\hline\bigstrut[t]
Laptop & LP & 62.39\% & 57.20\% & 59.68\%\\
Laptop & CNN & 76.31\% & 86.46\%  & 81.06\%\\
Laptop & CNN+LP & \bf 78.35\% & \bf 86.72\%  & \bf 82.32\%\\
Restaurant & LP & 65.41\% & 60.50\% & 62.86\%\\
Restaurant & CNN & 85.01\% & 87.42\% & 86.20\%\\
Restaurant & CNN+LP & \bf 86.10\% & \bf 88.27\% & \bf 87.17\%
\bigstrut[b]\\\hline%\hline
\end{tabular}
\caption{Impact of Sentic Patterns on the SemEval 2014 dataset}
\label{semeval2}
\end{table}

\begin{table}[h]
\centering%\small
\begin{tabular}{llccc}
\hline\bigstrut
Domain & Framework & Recall & Precision & F-Score\\
\hline\bigstrut[t]

Laptop & ZW & 66.51\% & 84.80\%  & 74.55\%\\

Laptop & CNN+LP & \bf 78.35\% & \bf 86.72\%  & \bf 82.32\%\\

Restaurant & ZW & 82.72\% & 85.35\% & 84.01\%\\

Restaurant & CNN+LP & \bf 86.10\% & \bf 88.27\% & \bf 87.17\%
\bigstrut[b]\\\hline%\hline
\end{tabular}
\caption{Comparison with the state of the art. ZW stands for \cite{zhiqiang2014dlirec}; LP stands for Sentic Patterns.}
\label{semeval3}
\end{table}

\makefigure{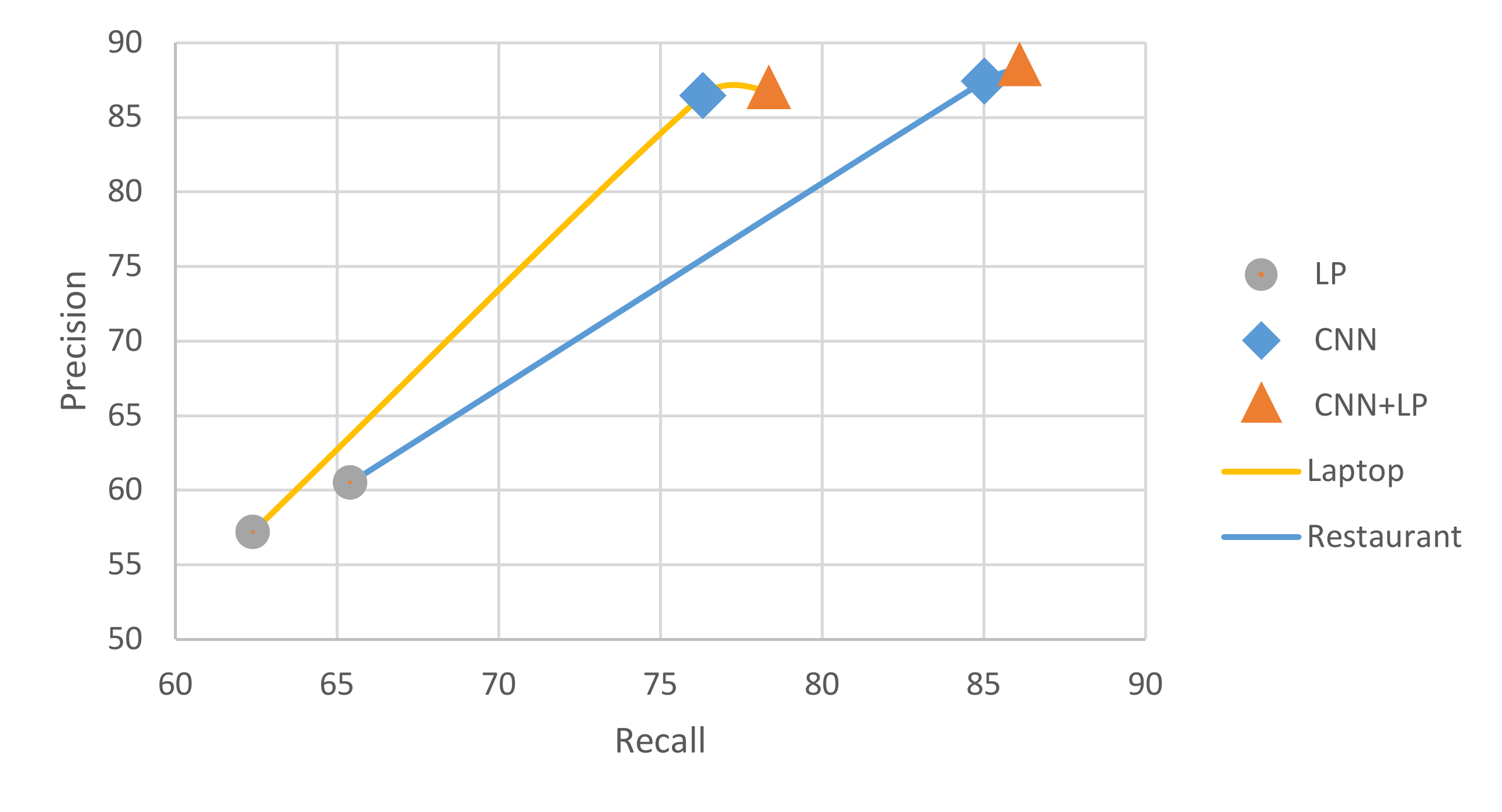}{Comparison between the performance of CNN, CNN-LP and LP.}

As to the LPs, the removal of stop-words, Rule~1, and Rule~3 were most beneficial. Figure \ref{fig:cnnlp} shows a visualization for the Table \ref{semeval2}. 
Table \ref{semeval3} and Figure \ref{fig:soa} shows the comparison between the proposed method and the state of the art on the Semeval dataset.
It is noted that about 36.55\% aspect terms present in the laptop domain corpus are phrase and restaurant corpus consists of 24.56\% aspect terms. The performance of detecting aspect phrases are lower than single word aspect tokens in both domains. This shows that the sequential tagging is indeed a tough task to do. Lack of sufficient training data for aspect phrases is also one of the reasons to get lower accuracy in this case. 

In particular, we got 79.20\% and 83.55\% F-score to detect aspect phrases in laptop and restaurant domain respectively. We observed some cases where only 1 term in an aspect phrase is detected as aspect term. In those cases Rule 4 of the LPs helped to correctly detect the aspect phrases.
We also carried out experiments on the aspect dataset originally developed by \cite{qiu2011opinion}. This is to date the largest comprehensive aspect-based sentiment analysis dataset.
The best accuracy on this dataset was obtained when word embedding features were used together with the POS features. This shows that while the word embedding features are most useful, the POS feature also plays a major role in aspect extraction. 

\begin{table}[t!]
\centering%\small
\label{bingliu2}
\begin{tabular}{llccc}
\hline\bigstrut
Domain & Classifiers & Precision & Recall & F-Score\\
\hline\bigstrut[t]
Canon & WE & 82.74\% & 75.15\%  & 78.76\%\\
Canon & WE+POS & \bf 85.42\% & \bf 77.21\%  & \bf 81.10\%\\
Nikon & WE & 73.19\% & 79.27\% & 76.10\%\\
Nikon & WE+POS & \bf 77.65\% & \bf 82.30\% & \bf 79.90\%\\
DVD & WE & 84.41\% & 77.26\% & 80.67\%\\
DVD & WE+POS & \bf 85.48\% & \bf 79.25\% & \bf 82.24\%\\
Mp3 & WE & 87.35\% & 81.23\% & 84.17\%\\
Mp3 & WE+POS & \bf 89.40\% & \bf 83.77\% & \bf 86.49\%\\
Cellphone & WE & 86.01\% & 81.32\% & 83.59\%\\
Cellphone & WE+POS & \bf 90.15\% & \bf 83.47\% & \bf 86.68\%
\bigstrut[b]\\\hline%\hline
\end{tabular}
\caption{Impact of the POS feature on the dataset by \cite{qiu2011opinion}}
\end{table}

As on the SemEval dataset, LPs together with CNN increased the overall accuracy. However, LPs have performed much better on this dataset than on the SemEval dataset. This supports the observation made previously \cite{qiu2011opinion} that on this dataset LPs are more useful. One of the possible reasons for this is that most of the sentences in this dataset are grammatically correct and contain only one aspect term. Here we combined LPs and a CNN to achieve even better results than the approach of by \cite{qiu2011opinion} based only on LPs. Our experimental results showed that this ensemble algorithm (CNN+LP) can better understand the semantics of the text than \cite{qiu2011opinion}'s pure LP-based algorithm, and thus extracts more salient aspect terms. Table \ref{bingliu3} and Figure \ref{fig:cnnlpdata} shows the performance and comparisons of different frameworks.

\makefigure{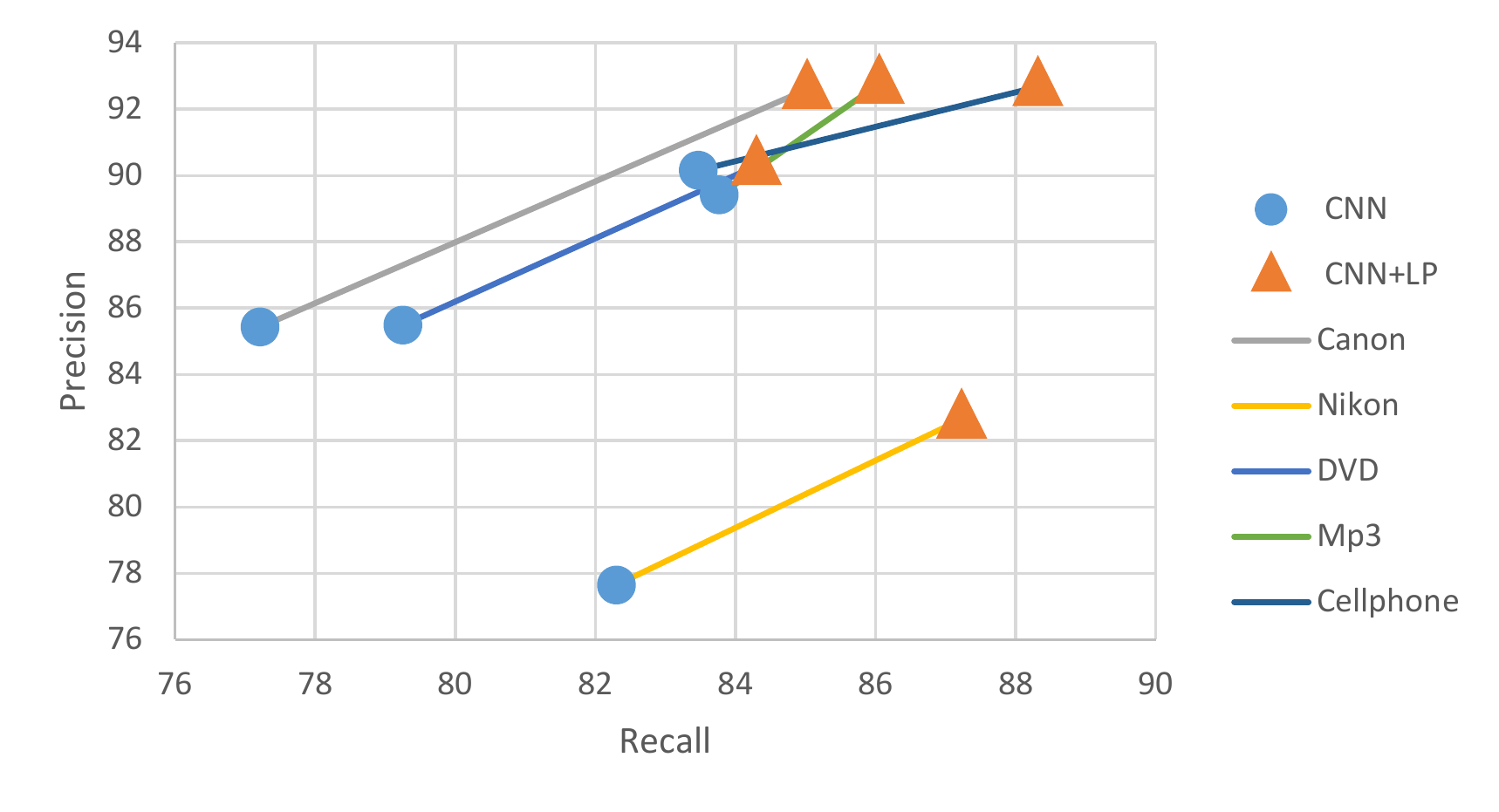}{Comparison between the performance of CNN, CNN-LP and LP.}

\begin{table}[t!]
\centering%\small
\begin{tabular}{llccc}
\hline\bigstrut
Domain & Classifiers & Precision & Recall & F-Score\\
\hline\bigstrut[t]
Canon & CNN & 85.42\% & 77.21\%  & 81.10\%\\
Canon & CNN+LP & \bf 92.59\% & \bf 85.02\%  & \bf 88.64\%\\
Nikon & CNN & 77.65\% & 82.30\% & 79.90\%\\
Nikon & CNN+LP & \bf 82.65\% & \bf 87.23\% & \bf 84.87\%\\
DVD & CNN & 85.48\% & 79.25\% & 82.24\%\\
DVD & CNN+LP & \bf 90.29\% & \bf 84.30\% & \bf 87.19\%\\
Mp3 & CNN & 89.40\% & 83.77\% & 86.49\%\\
Mp3 & CNN+LP & \bf 92.75\% & \bf 86.05\% & \bf 89.27\%\\
Cellphone & CNN & 90.15\% & 83.47\% & 86.68\%\\
Cellphone & CNN+LP & \bf 92.67\% & \bf 88.32\% & \bf 90.44\%
\bigstrut[b]\\\hline%\hline
\end{tabular}
\caption{Impact of Sentic Patterns on the dataset by \cite{qiu2011opinion}}
\label{bingliu3}
\end{table}

\makefigure{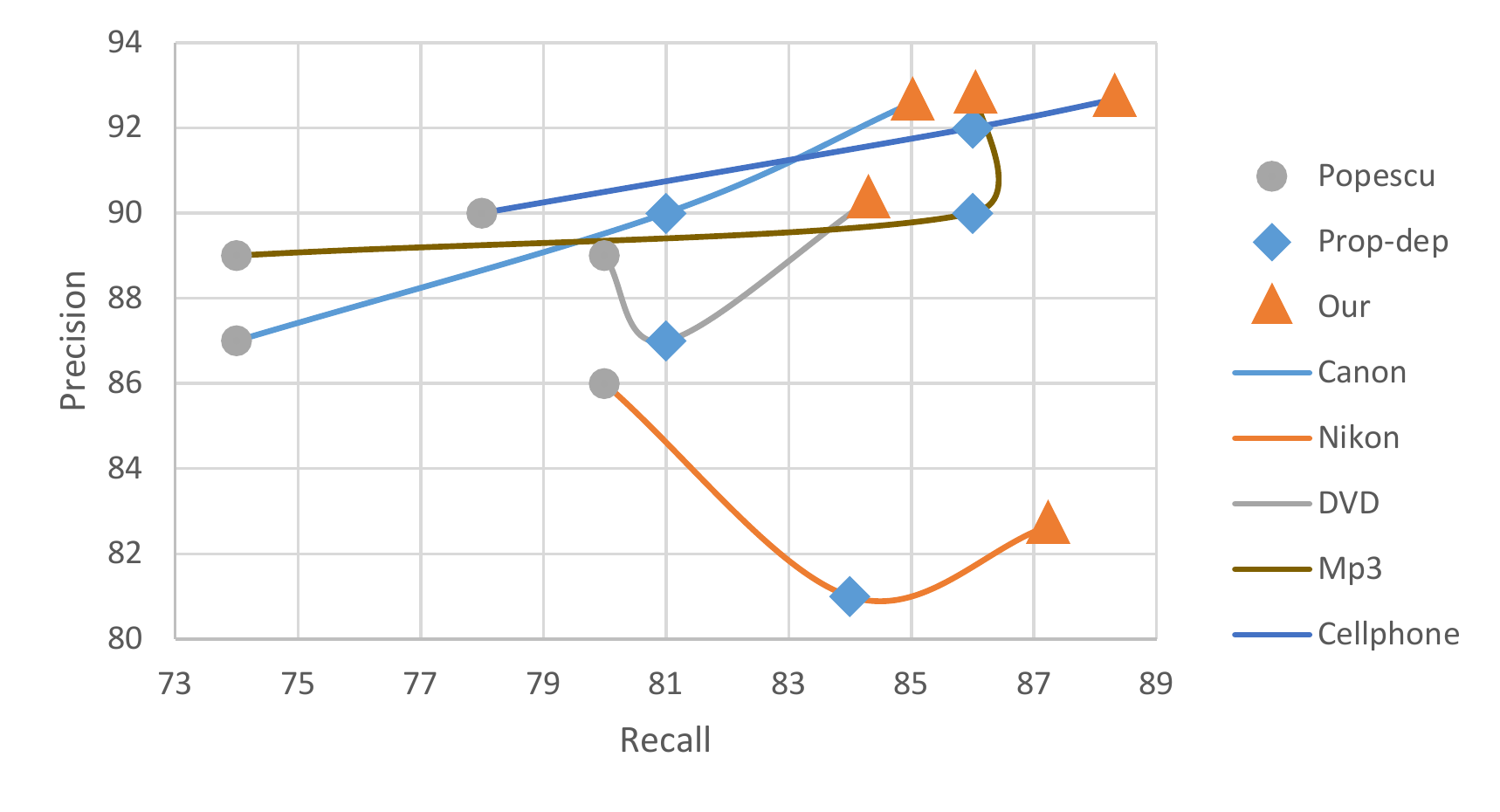}{Comparison of the performance with the state of the art on Bing Liu dataset.}

Figure \ref{fig:liub} compares the proposed method with the state of the art.
We believe that there are two key reasons for our framework to outperform state-of-the-art approaches. 
First, a deep CNN, which is non-linear in nature, better fits the data than linear models such as CRF. 
Second, the pre-trained word embedding features help our framework to outperform state-of-the-art methods that do not use word embeddings. 
The main advantage of our framework is that it does not need any feature engineering. This minimizes development cost and time. 

\section{Key Applications}

Subjectivity detection can prevent the sentiment classifier from considering irrelevant or potentially misleading text. This is particularly useful in multi-perspective question answering summarization systems that need to summarize different opinions and perspectives and present multiple answers to the user based on opinions derived from different sources. It is also useful to analysts in government, commercial and political domains who need to determine the response of the people to different crisis events. After filtering of subjective sentences, aspect mining can be used to provide clearer visibility into the emotions of people by connecting different polarities to the corresponding target attribute. 

\section{Conclusion}
\label{end}
In this chapter, we tackled the two basic tasks of sentiment analysis in social media: subjectivity detection and aspect extraction. We used an ensemble of deep learning and linguistics to collect opinionated information and, hence, perform fine-grained (aspect-based) sentiment analysis.
In particular, we proposed a Bayesian deep convolutional belief network to classify a sequence of sentences as either subjective or objective and used a convolutional neural network for aspect extraction. Coupled with some linguistic rules, this ensemble approach gave a significant improvement in performance over state-of-the-art techniques and paved the way for a more multifaceted (i.e., covering more NLP subtasks) and multidisciplinary (i.e., integrating techniques from linguistics and other disciplines) approach to the complex problem of sentiment analysis.

\section{Future Directions}

In the future we will try to visualize the hierarchies of features learned via deep learning. We can also consider fusion 
with other modalities such as YouTube videos. 

\section{Acknowledgement}
This work was funded by Complexity Institute, Nanyang Technological University. 

\section{Cross References}
Sentiment Quantification of User-Generated Content, 110170
Semantic Sentiment Analysis of Twitter Data, 110167
Twitter Microblog Sentiment Analysis, 265

\bibliographystyle{abbrv}
\bibliography{sigproc}

\end{document}